\documentclass[dvipsnames,sigconf=true, nonacm=false, review=False, anonymous = false,]{acmart}
\usepackage{graphicx} % Required for inserting images
\usepackage{xspace}
\usepackage{amsmath} % For align environment
\usepackage{subcaption}
\usepackage{xcolor}
\usepackage{url}

\copyrightyear{2026}
\acmYear{2026}
\setcopyright{cc}
\setcctype{by}
\acmConference[GECCO '26]{Genetic and Evolutionary Computation Conference}{July 13--17, 2026}{San Jose, Costa Rica}
\acmBooktitle{Genetic and Evolutionary Computation Conference (GECCO '26), July 13--17, 2026, San Jose, Costa Rica}
\acmDOI{10.1145/3795095.3805145}
\acmISBN{979-8-4007-2487-9/2026/07}
\newcommand{\SBS}{\textit{SBS}\xspace}
\newcommand{\VBS}{\textit{VBS}\xspace}
\newcommand{\SBP}{\textit{SBP*}\xspace}
\newcommand{\VBP}{\textit{VBP}\xspace}
\newcommand{\kSBS}[1]{\mbox{\textit{#1-SBS}}\xspace}
\newcommand{\kSBP}[2]{\mbox{\textit{#1-SBP*-w$^{#2}$}}\xspace}
\newcommand{\kOUR}[2]{\mbox{\textit{#1-SBP-w$^{#2}$}}\xspace}

\newcommand{\F}{\ensuremath{\mathcal{F}}\xspace}  %% Function set
\newcommand{\A}{\ensuremath{\mathcal{A}}\xspace}  %% Algorithm set
\newcommand{\B}{\ensuremath{\mathcal{B}}\xspace}  %% Budgets set
\newcommand{\E}{\ensuremath{\mathcal{E}}\xspace}  %% Precission set
\newcommand{\MS}{\ensuremath{\mathcal{S}}\xspace}  %% Solution multi-set
\newcommand{\fopt}{\ensuremath{f^{\text{opt}}}\xspace} %% Function optimum value
\newcommand{\fbest}{\ensuremath{f^{\text{best}}}\xspace} %% Function best
\newcommand{\af}{\ensuremath{\widehat{EAF}\xspace}} %%  Attainment Function
\newcommand{\perf}{\ensuremath{\mathcal{J}\xspace}} %% portfolio Performance
\newcommand{\pair}[2]{\ensuremath{\langle #1,\ #2 \rangle}}

\newcommand{\featfunc}{\phi}

\newcommand{\dist}{\delta}
\title[Similarity-based Portfolio Construction for Black-box Optimization]{Similarity-based Portfolio Construction \\for Black-box Optimization}

\author{Catalin-Viorel Dinu}
\affiliation{%
  \institution{Sorbonne Université, CNRS, LIP6}
  \city{Paris}
  \country{France}
}
\email{Catalin-Viorel.Dinu@lip6.fr}

\author{Diederick Vermetten}
\affiliation{%
  \institution{Sorbonne Université, CNRS, LIP6}
  \city{Paris}
  \country{France}
}
\email{Diederick.Vermetten@lip6.fr}

\author{Carola Doerr}
\affiliation{%
  \institution{Sorbonne Université, CNRS, LIP6}
  \city{Paris}
  \country{France}
}
\email{Carola.Doerr@lip6.fr}

\begin{document}
\begin{abstract}

In black-box optimization, a central question is which algorithm to use to solve a given, previously unseen, problem. Selecting a single algorithm, however, entails inherent risks: inaccuracies in the selector may lead to poor choices, and even well-performing algorithms with high variance can yield unsatisfactory results in a single run. A natural remedy is to split the evaluation budget across multiple runs of potentially different algorithms. Such sequential algorithm portfolios benefit from variance reduction and complementarities between algorithms, often outperforming approaches that allocate the entire budget to a single solver.

While effective portfolios can be constructed post-hoc, transferring this idea to the algorithm selection setting is non-trivial. We show that a naïve  portfolio constructed over the full training set already outperforms the strongest traditional baseline, the virtual best solver. We then propose a simple yet effective $k$-nearest-neighbor–based finetuning approach to construct portfolios tailored to unseen instances, yielding further improvements and highlighting the effectiveness of portfolio selection in fixed-budget black-box optimization.

\end{abstract}

\maketitle

\section{Introduction}

In the context of black-box optimization, a lot of research has been done into algorithms that are able to overcome specific challenges in optimization landscapes. This has led to the development of a wide range of optimization algorithms, with their own relative strengths and weaknesses. Naturally, this brings the question of how to choose the right optimizer for an unseen problem to the foreground. This algorithm selection problem~\cite{rice1976algorithm} has increasingly been combined with modern machine learning techniques, leading to promising results on a variety of testbeds~\cite{kerschke2019automated}. 

In the common algorithm selection context, the goal is to select an optimizer from a given set to be run on the unseen instance. Problems are generally represented by a set of features (e.g. exploratory landscape features~\cite{mersmann2011exploratory}), which are then combined with classifier systems to select the most promising algorithm from the available set. However, these selection techniques are not perfect, so there is an inherent risk that a non-optimal algorithm is chosen. Additionally, the available algorithms are generally stochastic in nature, so even if the optimal selection is made, it might still result in worse than expected performance. 

In the algorithm selection context, the impact of algorithmic stochasticity can be large because we select only a single run of an optimizer. If the selected algorithm were run repeatedly, these effects would be less impactful. Even in a fixed-budget scenario, repeated shorter runs of a high-variance optimizer might be more beneficial than a single run using the full budget. Additionally, the issue of selecting the wrong algorithm, for example, when the classification model is not reliable, could be mitigated in part by dividing the budget between multiple optimizers. 

Combining these two aspects, we can reformulate the traditional algorithm selection problem into a budget allocation problem, where we aim to create a portfolio of algorithms rather than select a single candidate. Previous work~\cite{ours} has shown that such algorithm portfolios can be constructed to perform well across sets of problems. These portfolios are constructed post-hoc based on the performance trajectories of all candidate algorithms on the given problem set. While it is not trivial to translate this construction one-to-one to the algorithm selection context, we show in this paper that even the naïve  method of constructing a portfolio on the training set already leads to promising performance. 

In addition to showcasing the benefits of sequential algorithm portfolios, we develop a finetuning method based on $k$-nearest-neighbors to slightly improve over this baseline. By exploring the impact of the used features, as well as the specifics of the neighborhood determination and portfolio construction settings, we showcase the potential of budget allocation as a more fine-grained formulation for algorithm selection.

\section{Related Work}
\begin{figure*}[tbp]
    \centering
    \includegraphics[width=0.7\textwidth]{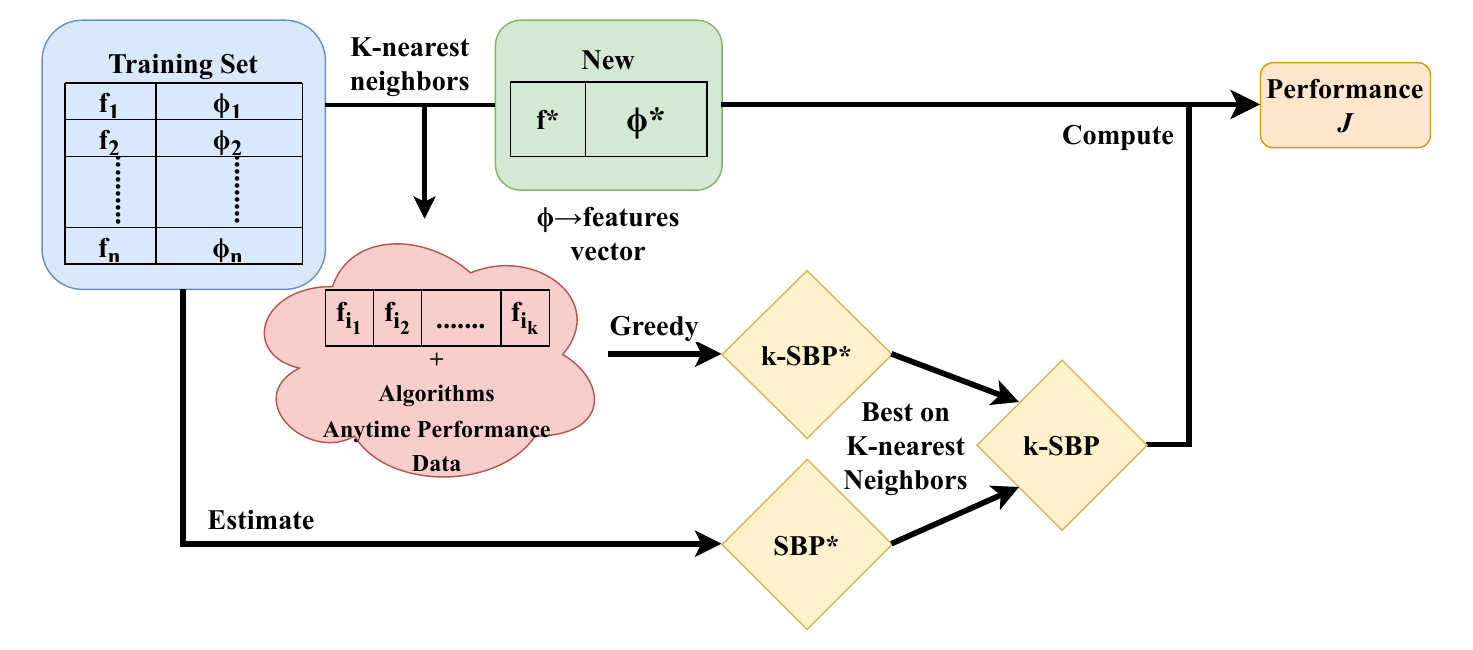}
    \caption{Methodology of the proposed approach. Given an unknown target function, a k-nearest-neighbor-based \textit{Single Best Portfolio} ($k$-\textit{SBP}$^{\ast}$) and the general \SBP are constructed. Both strategies are evaluated on the k-nearest neighboring functions, and the one with superior performance is selected as the final portfolio.}
    \label{fig:methodology}
\end{figure*}
Algorithm selection has been a topic of interest for a wide variety of problem types for a long time~\cite{rice1976algorithm}. The question of which method would be the most suitable to solve a new problem instance occurs naturally as soon as different methodologies with their own strengths and weaknesses are developed. Common algorithm selection methodologies frame the problem as a machine learning task, where a given set of instances and corresponding algorithm performances are known, and based on this data, a model should be created to predict the most performant algorithm for a set of unknown instances. In this context, the problems are generally represented by a set of features, based on which a machine learning model can be constructed~\cite{kerschke2019automated, smith2009cross}. 

The key reason for algorithm selection lies in the complementarity of the optimizers that can be chosen. However, selection is not the only way in which this type of complementarity can be exploited. Rather than building a selection model, one can also construct a portfolio consisting of complementary algorithms~\cite{leyton2003portfolio, guerri2004learning, gomes2001algorithm}. This is typically done in settings where parallel execution is possible, as in this context, there is no added cost to running multiple algorithms. However, even in settings where sequential execution is expected, the usage of algorithm portfolios can lead to improvements over sticking with a single algorithm. 

The construction of algorithm portfolios in a sequential evaluation context with a fixed total budget can be viewed as a budget allocation problem, where the total execution time, or number of function evaluations, can be divided across all candidate algorithms. In this setting, it can even be beneficial to allocate multiple repetitions of the same algorithm to benefit from variance reduction. Recent work in the black-box optimization context has shown that these types of algorithm portfolios can improve significantly over the static algorithm choices~\cite{ours}, motivating their use in an algorithm selection context. 

\section{Methodology}

This work can be viewed as an extension of the classical Algorithm Selection (AS) problem, in which the goal is no longer to identify a single best algorithm, but to construct an instance-specific algorithm portfolio. As in standard AS settings, we rely on a set of instance features that can be computed for both benchmark functions and previously unseen problem instances. These features provide a compact representation of problem characteristics and form the basis for relating new instances to past observations~\cite{cenikj2026survey}.

Whereas traditional AS methods typically use this information to select a single solver, our approach leverages it to guide portfolio construction. In particular, following the intuition of kNN-based AS methods~\cite{inproceedings}, we use distances in feature space to identify benchmark instances that are similar to the new problem. Rather than interpreting this neighborhood as a means to predict the best-performing algorithm, we view it as an estimate of the expected optimization behavior of the unseen instance. This estimate is then used to construct a portfolio of algorithms whose complementary strengths are well aligned with the inferred problem characteristics, allowing the method to balance robustness and specialization.

Figure~\ref{fig:methodology} provides a schematic overview of the proposed methodology. The approach starts from a finite training set $\F_d$ of black-box optimization problems, for which we assume access to anytime performance data of a predefined set of algorithms $\A$. To characterize and compare algorithm behavior throughout the optimization process, we employ the discretized empirical attainment function (EAF) as the anytime performance metric~\cite{grunert2001inferential, lopez2024using}:
\begin{equation*}
    \af_{f,\alpha}(b, \varepsilon) =
    \widehat{\mathbb{P}}\!\left[ \fbest_{\alpha}(b) - \fopt \leq \varepsilon \right],\, \forall \varepsilon \in \mathcal{E},\, \forall b \in \mathcal{B}.
\end{equation*}
which estimates the probability that algorithm $\alpha$ reaches a target solution quality $\varepsilon$ within a budget $b$ for a function $f$ to be optimized, where $\mathcal{E}$ denotes the set of considered thresholds and $\mathcal{B}$ denotes a set of equally spaced budgets.

Following previous work~\cite{ours}, a portfolio is defined as a multiset of algorithm--budget pairs,
\begin{equation*}
\MS = \left\{ (\alpha, b) \mid \alpha \in \A,\; b \in \B \right\},
\end{equation*}
which represents a sequential allocation of a fixed total budget~$T$. Portfolio performance is evaluated by aggregating attainment probabilities across problem instances and quality thresholds, thereby capturing the combined effect of multiple algorithm runs.

Let $\F_d^k \subseteq \F_d$ denote the set of functions on which the portfolio is evaluated, and let $w \in \mathbb{R}^{k}$ be a weight vector satisfying $\sum_{i=1}^{k} w_i = 1$. The performance of a portfolio $\MS$ is defined as
\begin{equation*}
\perf(\F_d^k, \MS, \vec{w}) =
\frac{1}{k \lvert \E \rvert}
\sum_{i=1}^{k}
w_i
\sum_{\varepsilon \in \E}
\left(
1 - \prod_{(\alpha,b) \in \MS}
\left( 1 - \af_{f_i,\alpha}(b, \varepsilon) \right)
\right).
\end{equation*}
This metric estimates the average probability that the portfolio attains a given quality threshold across all considered thresholds and functions, while explicitly accounting for the sequential combination of algorithm runs within the portfolio.

\subsection{Greedy Portfolio Construction Method}
Portfolios are constructed using a greedy procedure introduced in~\cite{ours}  that iteratively adds algorithm–budget pairs. Starting from the empty portfolio, and until the budget is exhausted, at each iteration $t$ the portfolio is extended by:
\begin{equation*}
\MS_{t+1} = \MS_t \oplus \pair{\alpha}{b}.
\end{equation*}
where candidate pairs are selected by maximizing a penalized performance score of the form:
\begin{equation*}
\text{score}(t+1, \pair{\alpha}{b})
= \perf(\F_d^k, \MS_t \oplus \pair{\alpha}{b}, \vec{w}) - 0.1\left(\frac{b}{T}\right)^2,
\end{equation*}
The penalty term $(b/T)^2$ discourages allocating excessively large budgets to a single algorithm and thereby promotes budget splitting and algorithm diversity. This penalized greedy strategy was shown to perform well compared to several alternative portfolio construction schemes and is therefore adopted here.

\subsection{Algorithm Selection Baselines}
Using the training benchmark set $\mathcal{F}_d$, we define several reference methods following standard conventions from the Algorithm Selection (AS) literature~\cite{kerschke2019automated}, which serve as baselines in our study. The \emph{Single Best Solver} \SBS is defined as the algorithm that achieves the best average performance over all training instances in $\mathcal{F}_d$. 

The \emph{Virtual Best Solver} \VBS is an oracle that, for each individual function in, selects the best-performing algorithm. While the \VBS is not realizable in practice, it provides an upper bound on achievable performance. Together, \SBS and \VBS constitute the strongest classical baselines commonly used in AS studies.

Extending the notion of single best solvers to portfolios, the \emph{Single Best Portfolio (SBP)} is defined as the portfolio that achieves the best average performance over the training benchmark set $\mathcal{F}_d$. The \emph{Virtual Best Portfolio (\VBP)} is an oracle portfolio that, for each individual function, selects the best-performing portfolio. Analogous to \textit{SBS} and \VBS, these objects serve as conceptual reference points, with \VBP providing an upper bound on achievable performance.

Since computing the \textit{SBP} exactly over $\mathcal{F}_d$ is computationally infeasible, we approximate it using a resampling-based procedure. Specifically, we sample multiple subsets $\mathcal{F}_d^m \subseteq \mathcal{F}_d$ of size $m$. For each sampled subset, we construct a portfolio using the greedy portfolio construction method described above, optimizing the performance metric restricted to that subset. Each candidate portfolio is then evaluated on the full training benchmark set $\mathcal{F}_d$. The portfolio achieving the highest average performance is selected as an approximation of the global \textit{SBP} and is denoted by \SBP. This globally optimized portfolio serves as the starting point of our method.

\subsection{Similarity-based Selection}
Similar to how proposed methods in the Algorithm Selection (AS) literature aim to reduce the performance gap between \SBS and \VBS, our goal is to narrow the gap between \textit{SBP} and \VBP by incorporating instance-specific information. While \SBS ignores instance characteristics entirely, the \emph{Single Best Portfolio} (\textit{SBP}) suffers from a similar limitation, as it represents a globally optimized solution that may be suboptimal for individual instances. In highly diverse benchmark scenarios, such global strategies may fail to be well suited for any particular instance.

To address this limitation, we adopt a traditional $k$-nearest neighbors (kNN) approach to construct portfolios tailored to previously unseen instances. Given a new function, we compute its feature representation ($\phi^{\ast}$) and identify a neighborhood $\mathcal{F}_d^k \subseteq \mathcal{F}_d$ consisting of the $k$ most similar training functions to $f^\ast$ according to the chosen similarity metric. Using this neighborhood, we construct a locally optimized portfolio by applying the same greedy portfolio construction procedure as before, but restricting performance evaluation to the functions in $\mathcal{F}_d^k$. The resulting portfolio is denoted by $k$-\textit{SBP}$^\ast$.

The portfolios \SBP and $k$-\SBP capture complementary strategies. The former represents a globally robust portfolio optimized across the entire training benchmark set, while the latter reflects a portfolio optimized for a specific neighborhood in feature space. Since the greedy portfolio construction process may be sensitive to the choice of optimization set, we explicitly compare the performance of \SBP and $k$-\textit{SBP}$^\ast$ on the neighborhood $\mathcal{F}_d^k$. The portfolio achieving higher average performance on this neighborhood is selected and assigned to the unseen function $f^\ast$. We denote the resulting portfolio by $k$-\textit{SBP}.

Furthermore, the $k$-nearest neighbors may contribute with different weights when computing the performance metric used for portfolio construction. This allows additional fine-tuning of the neighborhood-based portfolio by emphasizing functions that are more similar to the target instance. We denote this weighted variant by $\kSBP{k}{}$.

In summary, \SBS, \VBS, \textit{SBP}, and \VBP serve as reference baselines; \SBP denotes a globally optimized portfolio estimated from training data; \kSBP{k}{} is a neighborhood-specific portfolio constructed using similarity information; and \kOUR{k}{} is the final portfolio produced by our method and applied to unseen optimization problems.

\section{Experiments}
\subsection{Experimental Setup}

\begin{table*}[htbp]
\centering
\setlength{\tabcolsep}{6pt}
\renewcommand{\arraystretch}{1.2}
\begin{tabular}{lcccccc}
\toprule
& \textbf{\kSBS{10}} 
& \textbf{\VBS} 
& \textbf{\SBP} 
& \textbf{\kSBP{10}{eq}} 
& \textbf{\kOUR{10}{eq}} 
& \textbf{\VBP} \\
\midrule
2D 
& $-0.19 \pm 5.08$ 
& $4.94 \pm 9.29$ 
& $14.19 \pm 19.02$ 
& $12.35 \pm 16.12$ 
& $\mathbf{14.23 \pm 18.34}$ 
& $15.68 \pm 18.91$ \\
5D 
& $-0.69 \pm 8.13$ 
& $3.87 \pm 10.12$ 
& $\mathbf{18.29 \pm 25.05}$ 
& $17.19 \pm 24.16$ 
& $17.96 \pm 24.53$ 
& $23.64 \pm 31.09$ \\
10D 
& $-0.02 \pm 6.55$ 
& $4.89 \pm 10.40$ 
& $14.56 \pm 22.57$ 
& $15.21 \pm 23.10$ 
& $\mathbf{15.21 \pm 22.99}$ 
& $21.81 \pm 34.26$ \\
\midrule
Overall 
& $-0.30 \pm 6.68$ 
& $4.57 \pm 9.93$ 
& $15.68 \pm 22.35$ 
& $14.92 \pm 21.45$ 
& $\mathbf{15.80 \pm 22.10}$ 
& $20.38 \pm 28.96$ \\
\bottomrule
\end{tabular}
\caption{Relative improvement (mean $\pm$ std) over \SBS obtained by \kSBS{k}, \VBS, \SBP, the kNN-based approach (\kSBP{10}{eq}), the proposed fine-tuned method (\kOUR{10}{eq}), and \VBP. Among all compared approaches, the fine-tuned method demonstrates superior performance compared to the standard \SBP (the proposed baseline portfolio method).}

\label{tab:improvement_against_SBS}
\end{table*}

A central challenge in evaluating algorithm selection and portfolio-based methods is achieving sufficient diversity in benchmark problem instances. To address this, we opt to base our experimental study on a function generator rather than a fixed-size benchmark suite, allowing us to systematically control and scale the number of problem instances considered.

Our main experiments are conducted on a large collection of many-affine BBOB benchmark functions (MA-BBOB~\cite{vermetten2025ma}), denoted by $\mathcal{F}_d$. In contrast to the standard BBOB suite, which provides a fixed and relatively small number of instances per function class, MA-BBOB enables the generation of a substantially larger set of problem instances, making it particularly suitable for algorithm selection and portfolio-based evaluation. The functions in $\mathcal{F}_d$ are constructed through affine transformations and weighted combinations of standard BBOB functions, yielding a controlled source of variability across instances.

We consider three problem dimensionalities, $d \in \{2, 5, 10\}$. For each one, we generate a set of $1000$ benchmark functions. This set is split into a training set of $900$ functions, which are assumed to be known during portfolio construction and selection, and a test set of $100$ previously unseen functions that are used exclusively for evaluation.

The algorithm set consists of four optimizers: CMA-ES \cite{hansen2016cma} (from ModularCMAES \cite{denobel2021}), Diagonal CMA-ES \cite{ros2008simple}, RCobyla \cite{powell1994direct}, and Differential Evolution\cite{storn1997differential}(from nevergrad \cite{nevergrad}). These algorithms were selected to provide a heterogeneous mix of search behaviors, ranging from covariance-based adaptation to derivative-free local and population-based global search.

All algorithms are evaluated under a fixed evaluation budget of $T = 2000 \cdot d$ function evaluations, ensuring a fair comparison. The same budget is used when computing all portfolios, thereby maintaining consistency between portfolio construction and evaluation.

As the feature representation, we employ Exploratory Landscape Analysis (ELA) features~\cite{mersmann2011exploratory}, which characterize optimization landscapes through sampling-based descriptors such as modality, ruggedness, and variable interactions. Following common practice in algorithm selection, we select a subset of ELA features that has been shown to be informative in prior work (e.g.,~\cite{long2022learning}). Each function $f \in \mathcal{F}_d$ is mapped to a standardized ELA feature vector of dimension $n_d$, with $75$, $63$, and $63$ features for problem dimensions $d = 2$, $5$, and $10$, respectively. All features are standardized prior to similarity computation. Specifically, each feature is linearly standardized to have zero mean and unit variance, using statistics computed exclusively on the training set.

Similarity between functions is measured using cosine similarity in the standardized feature space. For two functions $f_i$ and $f_j$, with feature vectors $\featfunc_i$ and $\featfunc_j$, the cosine similarity is defined as
\[
\dist\bigl(\featfunc_i, \featfunc_j\bigr)
= \frac{\featfunc_i^\top \featfunc_j}
{\|\featfunc_i\|_2 \, \|\featfunc_j\|_2}.
\]
This measure captures the angular similarity between feature vectors and is invariant to their magnitude, making it well suited for comparing standardized ELA representations.

Using the proposed framework, the first experiment evaluates a simplified instantiation of our method in which both the neighborhood size and the weighting scheme are fixed. Specifically, we set \(k = 10\) and assign uniform weights to the selected neighbors, i.e., \(w^{eq}_i = \frac{1}{k}\). This represents the default variant of the method. For the estimation of \SBP, we sample 50 times a subset of 10 training functions, and we evaluate the resulting portfolios on the whole training dataset to choose the one with the highest performance. 

To ensure the reproducibility of our work, we created a Zenodo repository containing the code used to generate, process, and visualize the results as shown in the following sections. This repository is available at~\cite{zenodo}. 
% expriemnts .... -> reproductibility \cite{zenodo}

\subsection{Main Results}
We start by analyzing the empirical performance of the proposed portfolio-based methods and comparing them to classical algorithm selection baselines across all considered problem dimensions.

Table~\ref{tab:improvement_against_SBS} reports the relative performance improvement of the considered methods with respect to the \SBS, computed as  $\frac{\perf - \perf_\SBS}{\perf_\SBS}$. We first observe that the Single Best Portfolio (\SBP) already improves upon the \SBS by more than $14\%$ across all considered dimensions. Similarly, the locally constructed portfolio \kSBP{10}{eq} achieves substantial gains over the \SBS.

Notably, the improvement obtained by \SBP is considerably larger than the gap between the \SBS and the \VBS, indicating that, in this benchmark setting, pure algorithm selection offers limited potential for improvement. In contrast, portfolio-based approaches are able to leverage budget splitting across multiple algorithms to achieve significantly larger gains, highlighting the importance of portfolio construction in this regime.

\begin{figure}[tbp]
    \centering
    \includegraphics[width=0.6\linewidth]{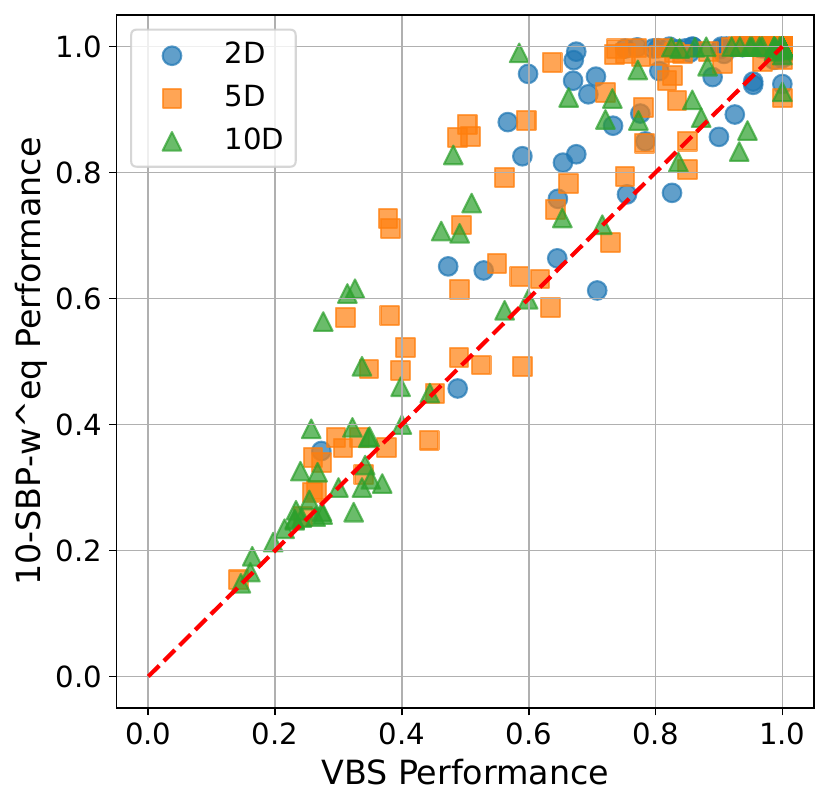}
    \caption{Pairwise performance comparison between \kOUR{10}{wq} (y-axis)  and \VBS (x-axis) across all benchmark instances. Each point corresponds to a single instance, illustrating that performance improvements are broadly distributed rather than confined to specific regions of the performance spectrum.} 
    \label{fig:VBS_VS_OURS}
\end{figure}

While \kSBP{10}{eq} slightly underperforms the global \SBP in the present experiments, this behavior is consistent with the relatively small \SBS-\VBS gap, which suggests limited benefits from fine-grained instance specialization. A similar trend is observed for portfolio-based methods, as reflected by the modest gap between \SBP and the \VBP. Furthermore, the comparatively weak performance of $\kSBS{10}$ mirrors the under-performance of \kSBP{10}{eq}, reinforcing the observation that local specialization becomes more beneficial as functional diversity increases. Indeed, this trend is already visible for higher problem dimensions, where the performance gap between global and local methods narrows.

The combined approach \kOUR{10}{eq}, which selects between the global and local portfolios based on their performance on the 10-nearest neighbors, consistently improves performance across all dimensions. Beyond the empirical gains, this behavior also highlights a limitation of the greedy portfolio construction procedure. In particular, there are cases in which the globally optimized portfolio \SBP outperforms the locally constructed $k\mathrm{-SBP}^\ast$ even when both are evaluated on the same local neighborhood. Under an ideal portfolio construction method, one would expect a portfolio optimized on the $k$-nearest neighbors to at least match the performance of the globally optimized portfolio on that neighborhood. The observed deviations therefore suggest that the greedy construction procedure does not always identify locally optimal portfolios, motivating the robustness gained by explicitly comparing global and local solutions.

Additional insights are provided by Figure~\ref{fig:VBS_VS_OURS}, which compares the performance of \kOUR{10}{wq} against the \VBS by plotting pairwise performance values across all instances. This comparison shows that performance improvements are broadly distributed rather than concentrated in specific regions of the performance spectrum. Notably, substantial gains are observed even on functions for which the \VBS already achieves strong performance.

Although minor performance degradations relative to the \VBS occur in a small number of cases, these losses remain limited in magnitude. Importantly, the proposed method avoids severe performance drops, indicating a favorable risk–reward trade-off. Overall, these findings suggest that the proposed approach delivers robust performance improvements while maintaining stability across a wide range of problem instances.

\subsection{\SBP vs \kSBP{k}{}}
\begin{figure}[tbp]
    \centering
    \includegraphics[width=0.7\linewidth]{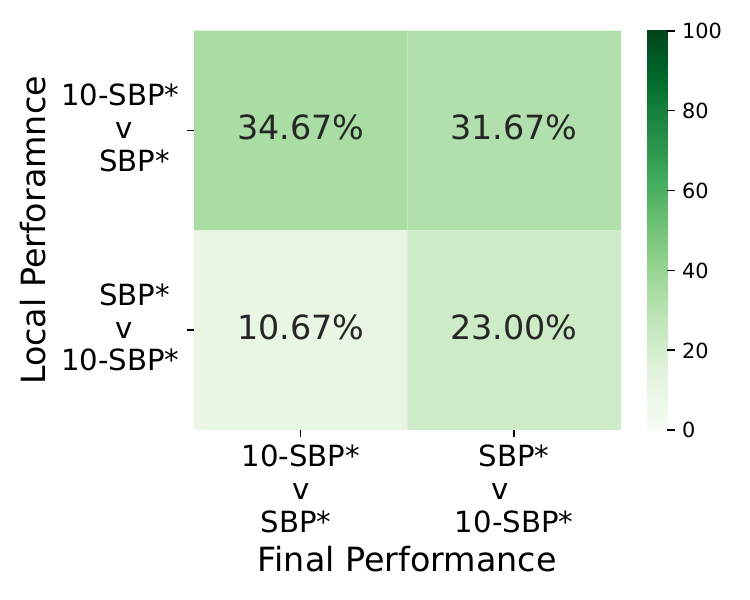}
    \caption{Distribution of the four possible cases defined by local and final performance when comparing \kOUR{10}{eq} and \SBP, aggregated over all problem dimensions. Each quadrant corresponds to a different agreement or mismatch between local neighborhood performance and final performance on the test instance.}
    \label{fig:percentage_each_method}
\end{figure}

\begin{figure*}
    \centering
    \includegraphics[width=\linewidth]{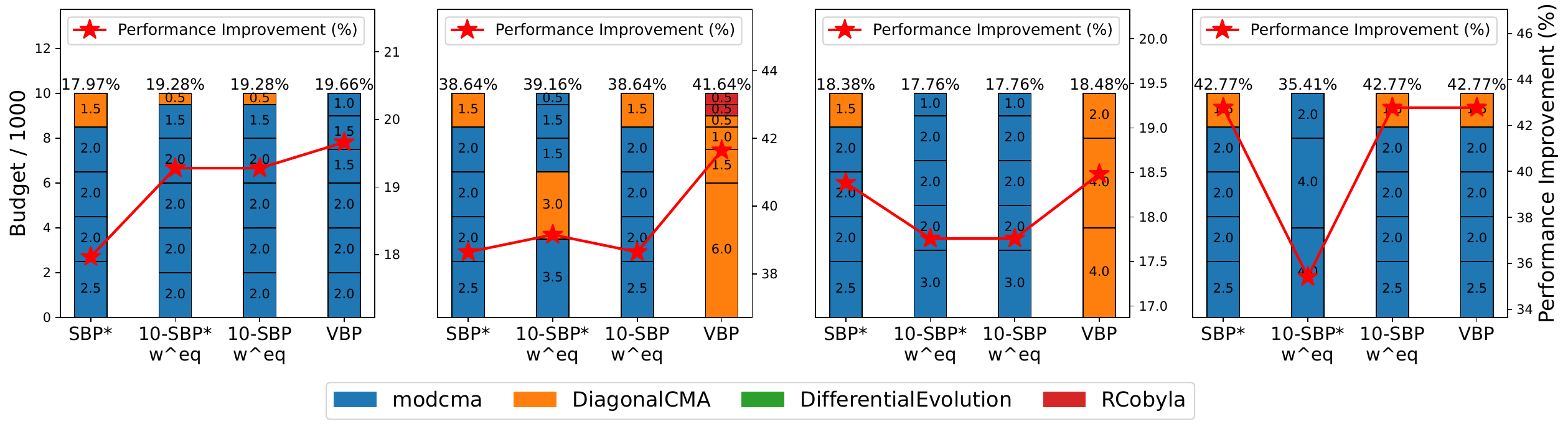}
    \caption{Representative portfolio compositions for the four local–final performance cases in the 5D setting. We compare portfolios constructed by \SBP, \kSBP{10}{eq}, \kOUR{10}{eq}, and \VBP. Colored bars represent the fraction of the total budget allocated to each algorithm (left y-axis), while the red line indicates the relative improvement over \SBS (right y-axis). The results show that portfolios are generally dominated by a small number of high-performing solvers across all cases.}
    \label{fig:portfolio_examples}
\end{figure*}

\begin{table*}[tbp]
\centering

\begin{tabular}{c
                c c
                c
                c
                c c
                c c}
\toprule

& \multicolumn{2}{c}{\textbf{\kSBS{10}}}
& \textbf{\VBS}
& \textbf{\SBP}
& \multicolumn{2}{c}{\textbf{\kSBP{10}{eq}}}
& \multicolumn{2}{c}{\textbf{\kOUR{10}{eq}}} \\
\cmidrule(lr){2-3}
\cmidrule(lr){6-7}
\cmidrule(lr){8-9}
& \textbf{ELA}
& \textbf{Final PERF}
& 
& 
& \textbf{ELA}
& \textbf{Final PERF}
& \textbf{ELA}
& \textbf{Final PERF} \\
\midrule
2D
& $-0.19 \pm 5.08$
& $4.83 \pm 9.27$
& $4.94 \pm 9.29$
& $14.19 \pm 19.02$
& $12.35 \pm 16.12$
& $12.71 \pm 15.35$
& $14.23 \pm 18.34$
& $\mathbf{14.36 \pm 18.66}$ \\
5D  
& $-0.69 \pm 8.13$
& $3.51 \pm 9.77$
& $3.87 \pm 10.12$
& $18.29 \pm 25.05$
& $17.19 \pm 24.16$
& $17.76 \pm 23.93$
& $17.96 \pm 24.53$
& $\mathbf{19.21 \pm 25.05}$ \\
10D 
& $-0.02 \pm 6.55$
& $4.70 \pm 10.47$
& $4.89 \pm 10.40$
& $14.56 \pm 22.57$
& $15.21 \pm 23.10$
& $16.31 \pm 21.99$
& $15.21 \pm 22.99$
& $\mathbf{16.65 \pm 22.79}$ \\
\midrule
\textbf{Overall}
& $-0.30 \pm 6.68$
& $4.35 \pm 9.82$
& $4.57 \pm 9.92$
& $15.68 \pm 22.33$
& $14.92 \pm 21.42$
& $15.59 \pm 20.77$
& $15.80 \pm 22.07$
& $\mathbf{16.74 \pm 22.31}$ \\
\bottomrule
\end{tabular}
\caption{Improvement (mean $\pm$ std) against \SBS comparison between the standard ELA feature representation and a performance-based latent feature space. The table highlights the impact of a more informative representation on \kSBS{10}, bringing its performance closer to the virtual best solver (\VBS).}
\label{tab:latent_space_comparison}
\end{table*}

To gain deeper insight into the differences between \kSBP{10}{eq} and \SBP, and how these differences affect performance, we distinguish four possible scenarios based on two binary comparisons. The first concerns local performance, evaluated on the $k$-nearest neighbors, and identifies which of \kSBP{10}{eq} or \SBP performs better on the selected neighborhood. The second concerns final performance, evaluated on the new test instance, and indicates which method ultimately achieves superior performance. Combining these two criteria yields four distinct cases, corresponding to agreements or mismatches between local and final performance.

Figure~\ref{fig:percentage_each_method} summarizes how often each of the four cases occurs. The upper-left quadrant corresponds to cases where \kSBP{10}{eq} outperforms \SBP both locally and in final performance, while the lower-right quadrant captures the opposite situation, where \SBP dominates both locally and globally. The remaining two quadrants represent mismatches between local and final performance, reflecting either over-specialization to the local neighborhood or successful adaptation despite weaker local performance.

Overall, $k$-nearest-neighbor–based portfolios outperform \SBP in approximately $45\%$ of the cases, compared to $55\%$ for \SBP. This relatively balanced split indicates that global portfolio optimization and local similarity-based adaptation contribute comparably to overall performance.

Since the final portfolio executed on a new function is selected based on local performance, cases in which \SBP outperforms $k$-SBP$^\ast$ on the $k$-nearest neighbors highlight a limitation of the greedy selection strategy underlying the locally optimized portfolio. In such situations, \SBP can be interpreted as providing a more effective locally optimized portfolio. Consistently, we observe that in approximately $58\%$ of the cases, the portfolio achieving the best local performance also yields superior final performance.

Given this categorization, Figure~\ref{fig:portfolio_examples} presents representative portfolio examples for each of the four cases, comparing portfolios obtained by \SBP, \kSBP{10}{eq}, \kOUR{10}{eq} and \VBP. Across all examples, a consistent pattern emerges: each portfolio is dominated by a small number of strong solvers, with CMA-ES appearing as a central component in every case. This suggests that portfolio selection primarily reallocates computational budget among high-performing solvers rather than replacing them entirely.

At the same time, these results highlight limitations related to the representativeness of the benchmark set. Even a comparatively heterogeneous collection such as MA-BBOB exhibits limited algorithmic diversity, making it challenging to assemble portfolios composed of solvers with comparable average performance and no clear dominant method. This lack of diversity constrains the observable potential of localized portfolio-based approaches and suggests that further gains may require benchmarks with more varied algorithmic strengths.

% \subsection{Final Performance as Features Space}
\subsection{More Informative Features Space}

In the literature, algorithm selection (AS) methods often suffer from limitations arising from the choice of function representation. Many such approaches assign algorithms to previously unseen functions based on a predefined feature space, which must therefore be sufficiently expressive to capture properties that relate problem instances to the expected performance of algorithms. Similarly, portfolio-based methods are affected by analogous limitations, as their effectiveness also depends on the expressiveness of the feature space used for instance characterization.

Motivated by this observation, we investigate how the proposed method behaves when operating in a richer latent feature space that contains more information about the expected performance of algorithms. For validation purposes, we consider an idealized setting in which we assume access to the final performance of all algorithms on all functions (both training and test) evaluated at budget~$T$. These performance values are then used to construct an alternative, performance-based feature representation.

Table~\ref{tab:latent_space_comparison} compares results obtained using the standard feature space (ELA features) with those achieved using this performance-based latent space. The first observation is that the more informative representation substantially improves the performance of \kSBS{10}, bringing it close to the virtual best solver \VBS, closing the gap between those two.

Similar to the results observed for kNN-based algorithm selection, examining the behavior of \kSBP{10}{eq} under the performance-based feature space reveals consistent improvements in portfolio performance. These gains also translate to the final portfolio method \kOUR{10}{eq}. In particular, the performance gap between \kSBP{10}{eq} and \SBP is effectively closed, and when propagated to the final selection strategy \kOUR{10}{eq}, this results in an additional average improvement of approximately $1\%$ relative to \SBS, compared to the standard \SBP.

While this idealized feature space can be interpreted as providing perfect information for algorithm selection methods, it represents only partial perfect information in the portfolio setting. Portfolio methods additionally differentiate between algorithms based on their performance throughout the optimization process, rather than solely on final performance at budget~$T$. This observation suggests that incorporating richer forms of performance information (such as intermediate or anytime behavior) could further enhance portfolio performance beyond what is achieved in the current idealized setting.

\begin{figure}
    \centering
    \includegraphics[width=0.7\linewidth]{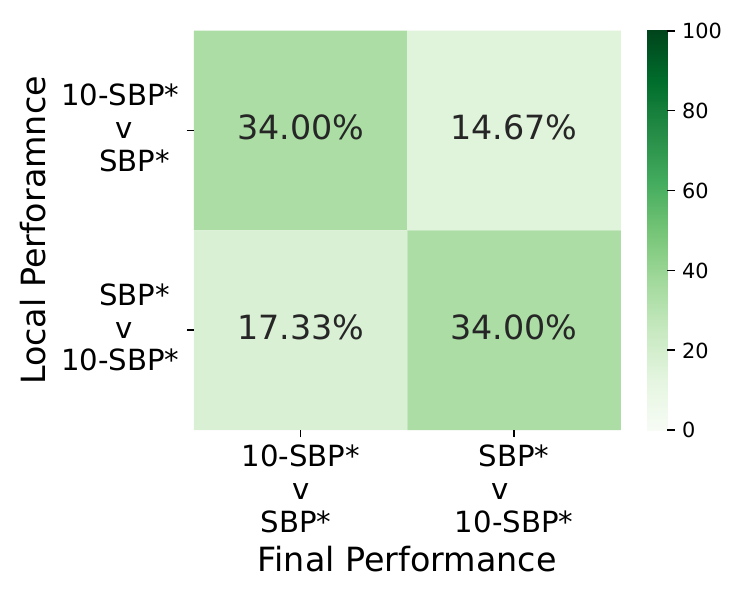}
    \caption{Distribution of the four possible cases defined by local and final performance when comparing \kOUR{10}{eq} and \SBP under the performance-based latent feature representation, aggregated over all problem dimensions.}
    \label{fig:percentage_each_method_perf}
\end{figure}

When examining the distribution of cases comparing \kSBP{10}{eq} and \SBP under the performance-based latent space (Figure~\ref{fig:percentage_each_method_perf}), we observe a clear shift in favor of the similarity-based portfolio selection method. In particular, \kSBP{10}{eq} outperforms \SBP in approximately $51.5\%$ of the cases, indicating that local adaptation based on $k$-nearest neighbors becomes beneficial in the majority of instances. At the same time, the fraction of cases in which selecting \kSBP{10}{eq} leads to a performance loss is further reduced compared to the standard feature space.

This shift underscores the strong dependence of the proposed method on the quality of the feature representation. A more informative latent space enables more reliable neighborhood identification, thereby improving the effectiveness of local portfolio adaptation. Overall, these results suggest that the $k$-nearest-neighbor component possesses substantial untapped potential, which can be more fully exploited when feature representations better capture algorithm performance characteristics.

\subsection{Neighborhood size}
\begin{figure}[htbp]
    \centering
    \includegraphics[width=\linewidth]{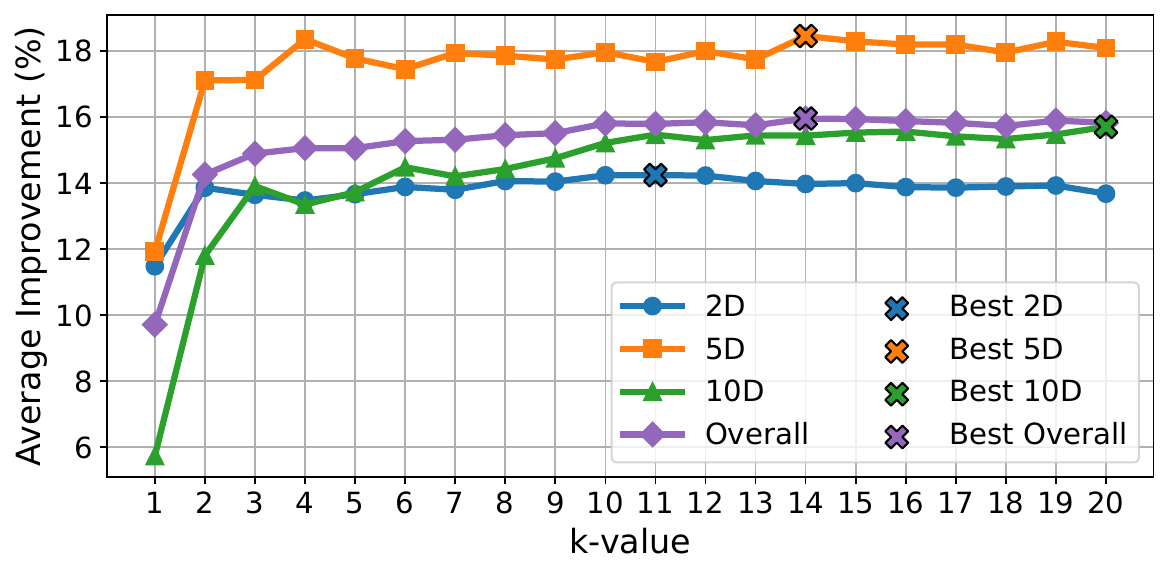}
    \caption{Relative improvement of \kOUR{k}{eq} over \SBS as a function of the neighborhood size, shown separately for each problem dimension and aggregated across all dimensions.}
    \label{fig:neighborhood}
\end{figure}

We now investigate how the neighborhood size affects the performance of the proposed method. Figure~\ref{fig:neighborhood} reports the relative improvement of \kOUR{k}{eq} over \SBS across the three individual problem dimensions, as well as the aggregated performance.

A first observation is that performance improvements are limited for small neighborhood sizes. This behavior is expected: when the neighborhood contains only a few similar functions, the resulting portfolio cannot reliably capture the variability of an unseen target function. In the extreme case where only a single neighbor is considered, the constructed portfolio may become overly specialized. Even minor mismatches between the expected and actual algorithm performance on a new function can then lead to suboptimal budget allocation or inappropriate solver selection, ultimately degrading performance.

As the neighborhood size increases, this effect is progressively mitigated. In our experiments, increasing the neighborhood size up to $k=10$ results in a clear stabilization of performance and consistently higher improvements relative to \SBS. Larger neighborhoods introduce greater functional diversity, which reduces the impact of feature–performance misalignment and yields portfolios that generalize more robustly to unseen functions.

Importantly, this observation also indicates that the minimum neighborhood size required for stable performance is closely linked to the quality of the feature representation. More informative and discriminative features would reduce misalignment between instance similarity and algorithm performance, thereby diminishing the need to rely on large neighborhoods to achieve robustness. In such settings, comparable performance could potentially be attained with smaller neighborhood sizes.

\subsection{Weight allocation in the function set}
While the neighborhood size controls which functions are considered similar to an unseen target problem, the similarity information itself can be further exploited to adapt how strongly each neighbor influences the constructed portfolio. In particular, differences in similarity can be incorporated through weighting schemes that emphasize closer neighbors while down-weighting more distant ones. We therefore investigate how different weighting strategies affect the performance of the similarity-based portfolio construction.

In the simplest setting, all functions in the neighborhood contribute equally to the performance metric used to construct $k$-SBP. More generally, the metric allows assigning different importance to individual neighbors, enabling functions that are closer in feature space to exert a stronger influence than those that are less similar.

Consider the $k$-nearest neighborhood of a function, sorted according to the similarity metric, where $d_i$ denotes the similarity of the $i$-th closest function and $d_{k+1}$ the similarity of the closest function not included in the neighborhood. We define the following weighting schemes:
\begin{align*}
w^{\mathrm{eq}}_i &= 1, \\
w^{\mathrm{soft}}_i &= e^{d_i}, \\
w^{\mathrm{diff}}_i &= d_i - d_{k+1}, \\
w^{\mathrm{log}}_i &= \ln\left(\frac{k+1}{2}\right) - \ln(i).
\end{align*}
The logarithmic weighting scheme follows the rank-based weighting used in CMA-ES \cite{hansen2016cma} and assigns higher importance to closer neighbors while gradually decreasing the influence of more distant ones. All weights are normalized such that $\sum_{i=1}^k w_i = 1$.

\begin{table}[htbp]
\centering
\setlength{\tabcolsep}{4pt}
\renewcommand{\arraystretch}{1.15}
\begin{tabular}{lcccc}
\toprule
\textbf{10-SBP}
& \textbf{w$^{eq}$}
& \textbf{w$^{soft}$}
& \textbf{w$^{log}$}
& \textbf{w$^{diff}$} \\
\midrule
2D  & $14.2 \pm 18.3$ & $\mathbf{14.3 \pm 18.3}$ & $13.9 \pm 18.4$ & $13.9 \pm 18.3$ \\
5D  & $18.0 \pm 24.5$ & $17.9 \pm 24.6$          & $\mathbf{18.5 \pm 25.9}$ & $18.4 \pm 26.2$ \\
10D & $15.2 \pm 23.0$ & $\mathbf{15.7 \pm 23.7}$ & $15.2 \pm 23.4$ & $15.0 \pm 23.0$ \\
\midrule
\textbf{Overall}
    & $15.8 \pm 22.1$ & $\mathbf{16.0 \pm 22.4}$ & $15.8 \pm 22.8$ & $15.8 \pm 22.7$ \\
\bottomrule
\end{tabular}
\caption{Performance (mean $\pm$ std) of \kOUR{10}{w} under different similarity-based weighting schemes, reported as relative improvement over \SBS. The results highlight the benefits of non-uniform weighting compared to uniform weighting.}
\label{tab:weights_comparison}
\end{table}

Table~\ref{tab:weights_comparison} reports the performance of \kOUR{10}{} under the different weighting schemes, measured as relative improvement over \SBS. The results show that incorporating non-uniform weights based on similarity consistently improves performance compared to uniform weighting. In particular, strategies that assign greater importance to closer neighbors enable the constructed portfolios to better reflect the expected behavior of the unseen target instance.

These findings indicate that the effectiveness of $k$-nearest-neighbor based portfolio construction depends not only on which neighbors are selected, but also on how their contributions are aggregated within the performance metric. By emphasizing the most similar functions, the portfolio construction becomes more sensitive to local structure in the feature space, leading to portfolios that are more specifically tailored to the characteristics of the target problem.

Moreover, the observed improvements further reduce the performance gap between \SBP and \VBP, suggesting that appropriate weighting schemes help the method move closer to idealized oracle behavior. This reinforces the close coupling between feature representation quality and weighting strategy in similarity-based portfolio selection, and suggests that jointly improving both aspects offers a promising direction for further performance gains.

\section{Conclusion}
In this work, we investigated similarity-based portfolio construction as a flexible alternative to traditional algorithm selection, with a particular focus on how neighborhood definition, weighting schemes, and feature representations influence performance. Our results show that portfolio-based approaches can serve as strong and robust baselines in settings where training and test instances are drawn from similar distributions. In such cases, constructing a high-quality portfolio on the training set can yield competitive performance without requiring additional feature computation at test time.

At the same time, our experiments highlight several important limitations and trade-offs. As in classical algorithm selection, the potential gains from portfolio fine-tuning are inherently bounded by the gap between the virtual best portfolio (\VBP) and the single best portfolio (\SBP), which was relatively small in our benchmark setting. When this gap is larger, more sophisticated selection and adaptation mechanisms are expected to become increasingly beneficial.

A key finding is the strong dependence of similarity-based methods on the quality and expressiveness of the feature representation. Comparisons with clustering-based approaches reveal that many of the challenges faced by machine-learning–based algorithm selection persist in the portfolio setting. While informed projection or transformation techniques~\cite{smith2023instance} could partially mitigate deficiencies in weak feature spaces, they cannot fully compensate for a lack of informative instance descriptors. Our analysis using an idealized, performance-based latent space shows that similarity-based portfolio selection has considerable untapped potential. When instance similarity aligns more closely with algorithm performance, local adaptation becomes both more effective and more reliable, closing much of the gap to oracle-like behavior.

Finally, our results demonstrate that the method offers substantial flexibility. Adjusting the neighborhood size and introducing non-uniform weighting schemes allows the portfolio construction to balance robustness against specialization. Larger neighborhoods with more uniform weights yield conservative but stable behavior, while smaller neighborhoods or more aggressive weighting enable finer adaptation to local structure at the cost of increased risk. This provides a practical mechanism to control the risk–reward trade-off of the selector.

\subsection{Future Work}
Several promising directions emerge from this study. First, an important avenue for future work is to investigate how the findings reported here transfer to settings characterized by a higher degree of algorithmic complementarity. In such scenarios, where different solvers excel on substantially different subsets of problem instances, portfolio construction is expected to play a more prominent role, potentially amplifying the benefits of similarity-based selection observed in this work.

It would also be worthwhile to investigate whether the creation of similarity-based algorithm portfolios might be beneficial in situations where the train and test distributions are more distinct. The stability gained from the portfolio creation might be useful in cases of distribution shifts~\cite{zhong2024sddobench} by providing a more robust algorithm that can further be fine-tuned if reliably similar instances are present. This fine-tuning could also be further adjusted to adapt dynamically to the situation, such as dynamically adjusting neighborhood sizes or weights based on prediction uncertainty, which could improve reliability over existing AS-based methods~\cite{rook2025efficient}. 

Third, portfolio construction could be combined with online or anytime algorithm selection techniques. As information about the current problem becomes available during optimization, such hybrid approaches could adapt the portfolio dynamically. Similarly, integrating warm-starting or information transfer between optimizers may further improve efficiency~\cite{kostovska2022per}, although care must be taken to preserve the variance-reduction benefits of independent runs. 

Finally, exploring alternative feature representations and distance metrics remains an important avenue for future work. Latent spaces that explicitly encode performance-related information, intermediate anytime behavior, or trajectory-based characteristics~\cite{jankovic2019adaptive} may substantially enhance similarity estimation and unlock further gains for portfolio-based selection methods.

\begin{acks}
We acknowledge funding by the European Union (ERC, ``dynaBBO'', grant no.~101125586).
% Views and opinions expressed are however those of the author(s) only and do not necessarily reflect those of the European Union or the European Research Council Executive Agency. Neither the European Union nor the granting authority can be held responsible for them. 
This research was also jointly funded by the French National Research Agency (ANR-23-CE23-0035) and the German Research Foundation (DFG; LI 2801/7-1), through project \textsc{Opt4DAC}.
\end{acks}

\bibliographystyle{ACM-Reference-Format}
\bibliography{bibliography} 
\end{document}